\documentclass{article}

\usepackage{amsmath}
\usepackage{amsthm}
\usepackage{amssymb}
\usepackage{graphicx}
\usepackage[preprint]{neurips_2025} 
\usepackage[utf8]{inputenc} 
\usepackage[T1]{fontenc}    
\usepackage{hyperref}       
\usepackage{url}            
\usepackage{booktabs}       
\usepackage{amsfonts}       
\usepackage{nicefrac}       
\usepackage{microtype}      
\usepackage{xcolor}         
\usepackage{multirow}

\usepackage{color, colortbl}
\usepackage{xcolor}
\usepackage{soul}

\title{ER-LoRA: Effective-Rank Guided Adaptation for Weather-Generalized Depth Estimation}

\author{
  Yan Weilong \quad Zhang Xin \quad Robby T. Tan \\
  National University of Singapore \\
  \texttt{yanweilong@u.nus.edu}
}

\begin{document}

\maketitle

\begin{abstract}

    Monocular depth estimation under adverse weather conditions (e.g., rain, fog, snow, and nighttime) remains highly challenging due to the lack of reliable ground truth and the difficulty of learning from unlabeled real-world data. Existing methods often rely on synthetic adverse data with pseudo-labels, which suffer from domain gaps, or employ self-supervised learning, which violates photometric assumptions in adverse scenarios.
    In this work, we propose to achieve weather-generalized depth estimation by Parameter-Efficient Fine-Tuning (PEFT) of Vision Foundation Models (VFMs), using only a small amount of high-visibility (normal) data. 
    While PEFT has shown strong performance in semantic tasks such as segmentation, it remains underexplored for geometry-centric tasks like depth estimation—especially in terms of balancing effective adaptation with the preservation of pretrained knowledge.
    To this end, we introduce the Selecting-Tuning-Maintaining (STM) strategy, which structurally decomposes the pretrained weights of VFMs based on two kinds of effective ranks (entropy-rank and stable-rank).
    In the tuning phase, we adaptively select the proper rank number as well as the task-aware singular directions for initialization, based on the entropy-rank and full-tuned weight; while in the maintaining stage, we enforce a principal direction regularization based on the stable-rank. This design guarantees flexible task adaptation while preserving the strong generalization capability of the pretrained VFM.
    Extensive experiments on four real-world benchmarks across diverse weather conditions demonstrate that STM not only outperforms existing PEFT methods and full fine-tuning but also surpasses methods trained with adverse synthetic data, and even the depth foundation model, under both supervised and self-supervised settings.

\end{abstract}

\section{Introduction}
\label{intro}

Monocular Depth Estimation (MDE) is a core task in computer vision, supporting a wide range of applications including robotics~\cite{yu2023udepth}, autonomous driving~\cite{zheng2024physical, deep-homo}, medical tasks~\cite{wang2025monopcc, pvchat}, reconstruction~\cite{ssnerf}, and scene understanding~\cite{schon2021mgnet, 3dswapping}.
Recent advances have largely centered on self-supervised MDE~\cite{2017left-right-consistency, godard2019digging,litemono, zhao2022monovit, bian2021ijcv_scdepth, 2023planedepth, 2021epcdepth,manydepth}, which avoid the need for ground-truth depth by enforcing photometric consistency across image pairs.
While effective in favorable environments, such methods often fail under adverse conditions (e.g., nighttime, fog, or rain), where the photometric assumption breaks down.
Another line of work focuses on building depth foundation models~\cite{depth_anything_v2, depthanything, marigold, gui2024depthfm, zoedepth, birkl2023midas, shao2023IEBins} via supervised learning on large-scale datasets, aiming to enhance cross-domain generalization.
However, their performance is limited by sparse and noisy annotations, which are common in outdoor adverse settings—where ground-truth depth (e.g., LiDAR) often covers only a fraction of the scene and suffers from artifacts like rain-induced reflections~\cite{gasperini_morbitzer2023md4all, nuscenes2019}.
Consequently, the inability to extract reliable supervision from unlabeled adverse data, combined with the inefficiency of relying on imperfect annotations, renders robust monocular depth estimation (RMDE) under diverse real-world conditions an unsolved and practically important problem.

To improve robustness under adverse conditions, several RMDE methods~\cite{wang2021RNW,diffusionadverse, gasperini_morbitzer2023md4all, Saunders_2023_ICCV, wang2023weatherdepth, diffusion_contrast, syn2real-depth} adopt unsupervised domain adaptation (UDA) frameworks to avoid reliance on real-world depth labels.
Early efforts~\cite{vankadari2020ADFA, wang2021RNW, spencer2020defeat, 23ICRA_Steps} target a single adverse scenario, such as nighttime, but often fail to generalize beyond their specific conditions.
Subsequent works~\cite{liu2021ADIDS, vankadari2023sun} attempt to bridge daytime and nighttime domains, yet struggle to handle the broader diversity of real-world environments.
More recent approaches~\cite{gasperini_morbitzer2023md4all, Saunders_2023_ICCV, diffusionadverse, wang2023weatherdepth, syn2real-depth} aim to improve generalization across diverse weather conditions by leveraging generative augmentation~\cite{CycleGAN2017, zheng_2020_forkgan, LDM, controlnet} or synthetic-to-real domain adaptation.
However, their reliance on synthetic data and limited exposure to diverse target domains hinders generalization to unseen real-world conditions.
These trends raise a critical question:
Is synthetic data truly necessary to achieve robust depth estimation under adverse conditions?

Currently, Vision Foundation Models (VFMs)~\cite{MAE, clip, EVA, oquab2023dinov2} have emerged as a powerful paradigm, demonstrating strong generalization across tasks and domains through large-scale self-supervised pretraining on natural images.
Recent works have dug into parameter-efficient fine-tuning (PEFT) of VFMs for tasks under challenging scenarios such as domain-generalized semantic segmentation~\cite{Rein, yun2024soma, bi2024fada, mfuser} and domain-generalized object detection~\cite{yun2024soma, chen2022adaptformer}, achieving promising results.
Motivated by their robustness and transferability, we naturally explore whether VFMs can be efficiently adapted to RMDE, particularly under unseen adverse conditions. 
However, existing PEFT methods \cite{hu2022lora, yun2024soma, miLoRA, task_specific_direction, Rein, chen2022adaptformer} are mainly designed for semantic tasks due to the semantic nature of VFM.
What's more, they overlook the full singular spectrum, have not considered critical direction protection, and underutilize the rich directional information in the frozen weight matrices.
On the contrary, RMDE relies on the low-level features instead of semantic information, which leads to sub-optimal results by simply utilizing previous approaches.


In this work, we present ER-LoRA, a novel PEFT pipeline for robust monocular depth estimation, based on the guidance of effective rank.
We begin by analyzing the frozen weight matrices of VFMs~\cite{oquab2023dinov2} via singular value decomposition (SVD), revealing the distribution of singular vectors across different singular values; also, we find how these weights have changed through full fine-tuning, with the structure similarity between the frozen weight and the residual weight.
Based on these insights, we design a Selecting, Tuning, and Maintaining (STM) strategy, aiming to adapt the semantically biased representations of VFMs toward low-level geometric reasoning while retaining their inherent robustness and generalization.
Guided by the definition of effective ranks (e-ranks)~\cite{entropyrank, stablerank3, stablerank1, stablerank2}, we adaptively select proper rank numbers for different weight matrices, as well as finding the most task-relevant directions. This helps with better initialization.
In the tuning phase, we only focus on training the low-rank branch; in the maitaining phase, we apply a component regularization to minimally perturb critical directions.
Compared to prior PEFT methods, our approach provides a more flexible task adaptation while preserving the strong generalization capability of the pretrained VFM, which is helpful for adapting strong VFMs to tasks that are semantics-agnostic.

In our experiments, we conduct evaluations in both self-supervised and supervised settings for RMDE, through training only on daytime data, and evaluate through multiple real-world datasets \cite{nuscenes2019, robotcar, drivingstereo, cadc} of diverse adverse and normal conditions (e.g., daytime, nighttime, rain, fog, snow) in a zero-shot manner. The experimental results of both settings show that our method surpasses previous synthetic-data-based methods, PEFT approaches, full fine-tuning (FFT), and the MDE foundation model. To conclude, our contributions can be listed as follows:

\begin{itemize}
    \item To our knowledge, we are the first to propose learning weather-generalized RMDE via Parameter-Efficient Fine-Tuning of VFMs. We design ER-LoRA, a novel PEFT pipeline for RMDE, which eliminates the need for synthetic adverse data used in prior methods.

    \item We leverage the definition of effective ranks to propose a novel Selecting, Tuning, and Maintaining (STM) strategy for PEFT, which gives flexible tuning space while retaining the pre-trained generalization ability and robustness compared with existing methods.

    \item Experimental results under both self-supervised and supervised settings across various real-world datasets of diverse conditions demonstrate our method outperforms existing synthetic-data-based methods, PEFT methods, full fine-tuning and MDE foundation models, with an average AbsRel enhancement of 7.3\% with FFT, 6.4\% with PEFT methods.
\end{itemize}

\section{Related Work}
\label{related_work}

\noindent \textbf{Monocular Depth Estimation} aims to predict a dense depth map from a single image, and has mainly evolved along two paradigms: supervised and self-supervised methods. The supervised methods recently tries to build up fundamental MDE models by leveraging strong backbones, such as convolutional neural networks \cite{Eigen2014, yuan2022newcrfs, depth_resnet2016}, transformers \cite{li2022depthformer, DPT2021, transdepth}, and generative models \cite{ marigold, gui2024depthfm, fu2024geowizard, hu2024metric3d}; or digging into data enhancement and scaling up \cite{birkl2023midas, zoedepth, depth_anything_v2, depthanything} by mixing various datasets, from synthetic to real-world, and from labeled to unlabeled data. Also, some works explore the introduction of language prior into the MDE pipelines \cite{VPD, zeng2024wordepthvariationallanguageprior, ecodepth, languageguidance}, and some research \cite{DAR} designs a depth autogressive model based on recent success in AR models. 

In contrast, self-supervised methods do not need any ground truth for supervision, which is based on the assumption of photometric consistency \cite{2017left-right-consistency, 2017cvpr_egomotion}. \cite{2017cvpr_egomotion} first utilizes photometric loss in self-supervised depth, while the following works \cite{godard2019digging, sc_depthv3, bian2021ijcv_scdepth} design different strategies to solve problems with moving objects and scale ambiguity. Some approaches \cite{zhao2022monovit, litemono} focus on designing a more lightweight architecture to efficiently combine attention and CNN. 

\noindent \textbf{Robust Monocular Depth Estimation} in adverse weather conditions is challenging due to diverse types of degradation, and the inefficiency of learning from real-world data. Some approaches are tailored to a specific condition \cite{vankadari2020ADFA, wang2021RNW, zhao2022ITDFA, 23ICRA_Steps, DCLdepth} or daytime-nighttime conditions \cite{liu2021ADIDS, vankadari2023sun, spencer2020defeat}, either in a discriminative learning manner, utilizing generative models \cite{CycleGAN2017} for data augmentation, or exploring image enhancement to a certain condition. Recent methods~\cite{gasperini_morbitzer2023md4all, Saunders_2023_ICCV, diffusionadverse, wang2023weatherdepth, diffusion_contrast, syn2real-depth} address multi-condition RMDE by generating realistic adverse data via GANs or diffusion models~\cite{CycleGAN2017, zheng_2020_forkgan, LDM, controlnet}, or designing a synthetic-to-real adaptation strategy. However, they often rely on a costly data synthesis process as well as the synthetic quality, and struggle to generalize to unseen conditions.

\noindent \textbf{Parameter-efficient Fine-tuning (PEFT) of Vision Foundation Models (VFMs)} targets at fine-tuning a very small subset of parameters to adapt the large VFM to downstream tasks. Recent methods include: Low-Rank Adaptation \cite{hu2022lora, yun2024soma, miLoRA, liu2024dora, fan2025makeloragreatagain, meng2024pissa}, which inject low-rank tuning branches into the model, achieving minimal disruption to its core representational structure; Adapter-tuning approaches \cite{chen2022adaptformer, bi2024fada, set, Rein, mfuser}, which insert lightweight modules after each layer to refine the latent representations without modifying the original backbone; and Prompt-tuning methods \cite{lester2021powerscaleparameterefficientprompt, jia2022visualprompttuning}, introduce learnable tokens to the input of selected attention layers, enabling task-specific conditioning without modifying model weights. Many other domain generalization or adaptation tasks \cite{domain_human_pose, adaptive-domain-gen, heap} also consider this. However, previous PEFT methods for VFMs focus on semantic tasks, with limited exploration of low-level, geometry-centric applications. Given the semantic bias of VFMs, adapting them to tasks like RMDE requires weight tuning. To this end, we build on the low-rank adaptation paradigm and propose a novel PEFT strategy for robust depth estimation.

\section{Preliminary}
\label{preliminary}

\subsection{Self-supervised Learning in MDE}
The self-supervised MDE assumes there exists photometric consistency between consecutive frames of monocular videos \cite{2017cvpr_egomotion, 2017left-right-consistency}, which is usually held in daytime conditions of good visibility. In our self-supervised pipeline, we follow the common practice of Monodepth2 \cite{godard2019digging} to conduct self-supervised training.
Given a target frame $I_t$ and a source frame $I_{t'}$, a depth network $\Phi_D$ and a pose network $\Phi_P$ are trained simultaneously to minimize the objective of photometric error:
\begin{align}
  L_{\rm pe}&= \text{PE} (I_t, I_{t^{'}\rightarrow t}) \label{equ:pe_loss}, \\
  \rm{PE}(a,b) &= \frac{\alpha}{2}(1-\text{SSIM}(a,b)) + (1-\alpha) \|a-b\|_1, \label{pe}\\
 \quad 
 I_{t^{'}\rightarrow t} &= I_{t^{'}}\langle \text {Proj} (D_t, T_{t\rightarrow t^{'}}, K) \rangle, \label{project}
 \end{align} 
where \(\text{PE}\) denotes the photometric error, \(\alpha\) is a weighting factor, \(K\) represents known camera intrinsics, \(\langle \cdot \rangle\) denotes the sampling operator, and \(\text{Proj}()\) is the projection operation. $I_{t^{'}\rightarrow t}$ means projecting pixels from $I_{t^{'}}$ to $I_{t}$, and $D_t$ and $T_{t\rightarrow t^{'}}$ represent the output of $\Phi_D$ and $\Phi_P$, respectively. In our method, we follow the previous design of $\Phi_P$, but dig into PEFT of VFMs in $\Phi_D$.

\subsection{Low-rank Adaptation (LoRA)}
LoRA \cite{hu2022lora} assumes that the weight change of fine-tuning for pretrained weight matrices can be modeled within a low-rank structure. Assuming the pretrained weight matrix inside VFMs is denoted as $W_0 \in \mathbb{R}^{ m \times n}$, and the update matrix for $W_0$ can be defined as $\Delta W \in \mathbb{R}^{ m \times n}$. Then $\Delta W$ can be decomposed in a low-rank manner:
\begin{align}
    \Delta W = B A,
\end{align}
where $B \in \mathbb{R}^{ m \times r}$, $A \in \mathbb{R}^{r \times n}$, and $r$ is the intrinsic rank with $r \ll \min(m, n)$. Noted that $A$ is initialized with a uniform Kaiming distribution \cite{he2015delvingdeeprectifierssurpassing}, and $B$ is initialized with zeros. This guarantees that $\Delta W$ is set to the zero matrix initially. The final weight matrix $W^{'}$ is defined as:
\begin{align}
    W^{'} = W_0 + \Delta W = W_0 + BA.
\end{align}

Notably, this design requires updating only a quite small number of parameters compared to full fine-tuning (FFT), while introducing negligible computational overhead during inference.

\subsection{Singular Value Decomposition (SVD) and Effective Ranks (e-ranks) of Matrices}

The rank of a matrix $W$ can be obtained from its Singular Value Decomposition by the number of non-zero singular values.
Eckart–Young–Mirsky theorem \cite{Eckart_Young_1936} revealed that the top $r$ singular components can be summed up to represent the core of the matrix. 
However, in real-world scenarios, large matrices often have full rank, even though many of the components are dominated by noise or carry little useful information.
We go into detailed information through the singular spectrum of weight matrices, with Effective Rank, which is the real-valued extension of rank.

\noindent \textbf{Definition 1. Effective Rank from Entropy of Singular Values (entropy rank).} \textit{Assuming a matrix $W \in \mathbb{R}^{ m \times n}$, which can be decomposed via SVD into}
\begin{align}
\label{svd}
   W = U \Sigma V^T  = \sum_{i=1}^{K} u_i \sigma_i v_i^T,
\end{align}
where $K=\min(m, n)$, $u_i, \sigma_i, v_i$ correspond to the $i^{\rm{th}}$ left singular vector, singular value, and right singular vector. \cite{entropyrank} models $\sigma_i$ as a probability distribution that $p_i = \frac{\sigma_i^\gamma}{\sum_{j=1}^{K} \sigma_j^\gamma}$. Then, the entropy of such a distribution is obtained by $H = - \sum_{i=1}^{K} p_i \log p_i$, and the entropy rank is defined as
\begin{align}
\label{entropy_rank}
    Rank_{\text{en}}(W) \triangleq e^H = \exp\left( -\sum_{i=1}^{K} p_i \log p_i \right).
\end{align}

Notably, as discussed in \cite{entropyrank}, the entropy rank reflects the dispersion of information across singular directions--larger values correspond to more evenly distributed singular values.

\noindent \textbf{Definition 2. Effective Rank from Numerical Perspective of Singular Values (stable rank).} \textit{\cite{stablerank1, stablerank2, stablerank3} define the stable rank from the relative magnitudes through all singular values. Since we have $1 \geq \frac{\sigma_2}{\sigma_1} \geq \cdots \geq \frac{\sigma_K}{\sigma_1} \geq 0$, the stable rank is defined as }
\begin{align}
\label{stable_rank}
Rank_{\text{st}}(W) \triangleq \sum_{i=1}^{K} \frac{\sigma_i^\gamma}{\sigma_1^\gamma}.
\end{align}

The stable rank reflects the degree of concentration of energy in the leading singular directions.
A lower stable rank indicates that most energy is concentrated in a few directions.

\textbf{Lemma 1.} \textbf{Stable rank is upper bounded by entropy rank.} 
\textit{The stable rank is always less than or equal to the entropy rank, assuming that $\gamma = 1$: $Rank_{\rm{st}}(W) \leq Rank_{\rm{en}}(W)$. The detailed proof is provided in the Appendix.}

\section{Methodology}
\label{method}
\begin{figure*}[tb]
\centering
\includegraphics[width=1.\linewidth, height=.29\linewidth]{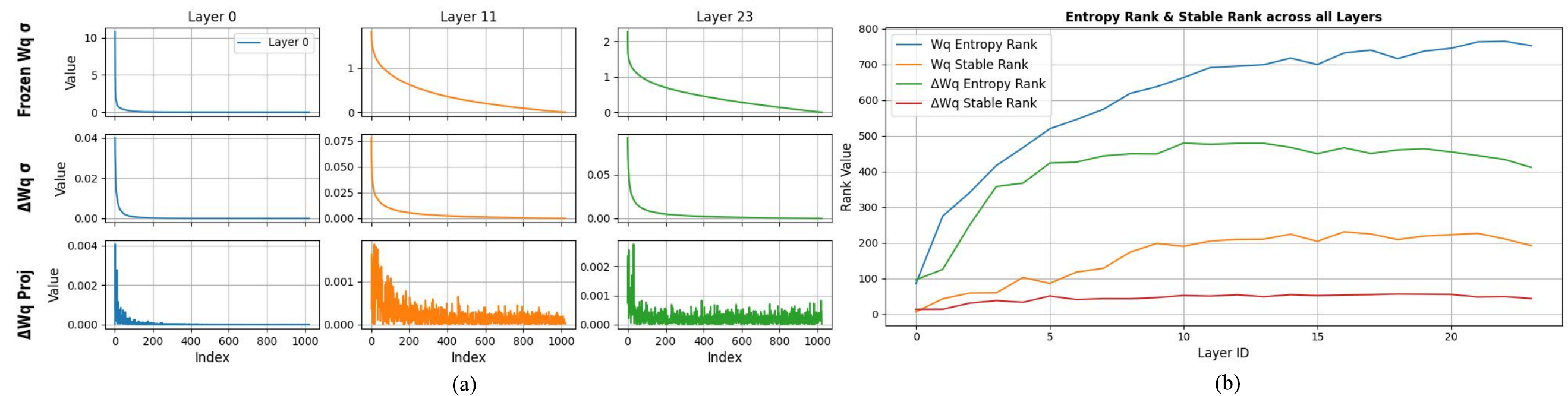}

\caption{(a) Row 1: Distribution of singular values within VFM's frozen weights. Row 2: Distribution of singular values within residual weights after full fine-tuning. Row 3: Projection values of projecting post-FFT weights onto singular components of frozen weights. (b) Illustration of the entropy rank and stable rank within frozen weights and residual weights, through all the layers.}
\label{fig:analysis}
\vspace{-0.3cm}
\end{figure*}


Fig. \ref{fig:pipeline} gives an overview of our proposed method. 
We start by analyzing the weight matrices inside VFMs via SVD through different layers in Fig. \ref{fig:analysis}, clearly studying how different weight matrices with different singular value distributions have been changed after full fine-tuning (FFT) in Sec. \ref{sec:analysis}.
With the observations, we propose the Selecting, Tuning, Maintaining (STM) Strategy for RMDE via PEFT of VFMs within each linear layer in Sec. \ref{STM strategy}, and our training pipeline in Sec.~\ref{sec:Training_Pipelines}.

\subsection{Weight Decomposition and Analysis inside VFMs}
\label{sec:analysis}

Inspired by \cite{yun2024soma}, we first decompose the weight matrices inside a frozen VFM (e.g., DINOv2 \cite{oquab2023dinov2}) into Eq. \ref{svd}. The singular values $\sigma$ across different layers can be seen in row $1$ of Fig. \ref{fig:analysis}(a), where we plot the spectra within $W_{\rm{q}}$, corresponding to the $0^{\text{th}},11^{\text{th}},23^{\text{th}}$ layers.
It is evident that deeper layers (e.g., Layer $23$) exhibit a more uniform singular value distribution compared to shallower ones (e.g., Layer $0$), as reflected by the slower decay curves in their spectra. 
This indicates that in shallow layers, very few components can stably provide most of the information, and it would be better not to disturb such a structure with large rank adaptation, retaining pretrained robustness. 

SoMA~\cite{yun2024soma} directly freezes the shallow layers, and uniformly samples the last $r$ singular components from each weight matrix for LoRA initialization.
Considering that RMDE is not strongly correlated with semantic understanding inside VFMs, we step further to find how weight matrices change from $W_{\rm{q}}$ to  $W'_{\rm{q}}$ after full fine-tuning (FFT) on a few daytime data. Here, the residual weight is defined as $\Delta W_{\rm{q}}=W'_{\rm{q}} - W_{\rm{q}}$. Similarly, we plot the singular values of $\Delta W_{\rm{q}}$ in row $2$ of Fig. \ref{fig:analysis}(a). As with $W_{\rm{q}}$, the spectra in shallower layers exhibit a much sharper decay. What's more, in order to find how the magnitude $\sigma_{i}$ of each direction $u_i v_i$ in Eq. \ref{svd} changes, we project $\Delta W_{\rm{q}}$ to  $u_i v_i$ via $|u_i^T \Delta W_{\rm{q}} v_i|$, as visualized in row $3$ of Fig. \ref{fig:analysis}(a)--the projection amplitudes in shallow layers decay rapidly and remain smooth, suggesting minimal disruption to their singular structure.  In contrast, deeper layers exhibit fluctuations across a broader range of directions, indicating that tuning perturbs a wide range of singular components. These observations lead to two important insights:
\textbf{(1) Tuning tends to induce higher-rank transformations in deeper weights, while structural changes in shallower layers remain relatively minor;
(2) In shallow layers, the perturbations are mostly concentrated in top singular components, whereas in deeper layers, the changes are more uniformly spread across the entire singular spectrum.}

Inspired by the above, we aim to quantify how much different layers have changed by introducing the computation of entropy rank in Eq. \ref{entropy_rank} and stable rank in Eq. \ref{stable_rank}, and plot them through all layers in Fig. \ref{fig:analysis}(b). Both the entropy and stable ranks of $W_{\rm{q}}$ increase with layer depth, indicating greater structural complexity in deeper layers. In contrast, $\Delta W_{\rm{q}}$ remains with a lower rank, though the rank also grows across layers, suggesting broader but still sparse perturbations. These trends motivate our Selecting-Tuning-Maintaining design: (1) \textbf{Selecting} the proper budget $r$ and task-aware directions for low-rank tuning; (2) \textbf{Tuning} only the low-rank branch to adapt to target tasks; (3) \textbf{Maintaining} the directions with concentration of energy from stable-rank.

\subsection{Selecting-Tuning-Maintaining (STM) Strategy}
\label{STM strategy}
\begin{figure*}[!tb]
\centering
\includegraphics[width=.98\linewidth, height=.46\linewidth]{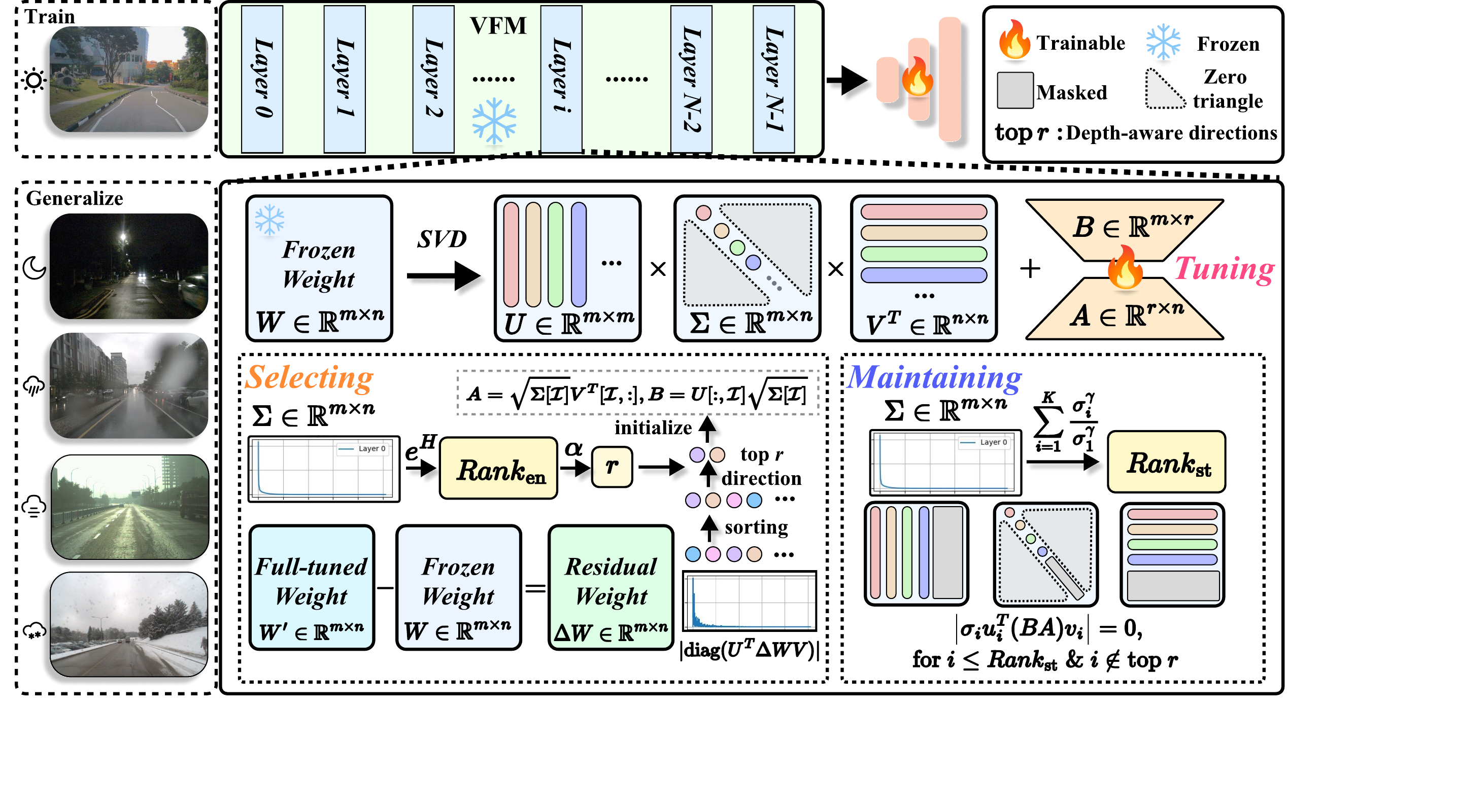}
\caption{The ER-LoRA pipeline. In the Selecting phase, we adaptively allocate proper rank numbers and task-aware directions for different weight matrices, based on the entropy-rank and full-tuned weight. In the Maintaining phase, the leading singular components are preserved via regularization.}
\label{fig:pipeline}
\end{figure*}

As shown in Fig. \ref{fig:pipeline}, the STM strategy starts by selecting proper objectives for low-rank adaptation initialization. A good initialization should include two keys: How significant to tune (ranks) and what way to tune (directions). Let the pretrained weight, full-tuned weight, and full-tuned residual weight be denoted as $W, W', \Delta W$, respectively. The low-rank tuning matrices are defined as $B \in \mathbb{R}^{ m \times r}$, $A \in \mathbb{R}^{r \times n}$. Then, based on the observations in Sec. \ref{sec:analysis}, we can set a simple but meaningful linear correlation between $r$ and the entropy rank $Rank_{\rm{en}}$ within $W$:
\begin{align}
\label{entropy_rank_to_rank}
    r = \alpha \times Rank_{\text{en}}(W),
\end{align}
where $\alpha$ serves as a scaling factor to control the magnitude of low-rank adaptation. This design allows flexible tuning capacity based on the singular spectrum of each pretrained weight matrix, distinguishing our approach from prior methods. Apart from the choice of ranks, we also explore which singular components are chosen for initialization. To achieve that, we extract the absolute projection value from the full-tuned residual weight on the directions of the pretrained weight:
\begin{align}
\label{projection}
    \mathbf{d} = |\operatorname{diag}(U^\top \Delta W V)| \in \mathbb{R}^{K \times 1},
\end{align}
where $\mathbf{d}$ contains the absolute diagonal values of the projection matrix. Different from \cite{task_specific_direction} that utilizes relative change from projection for semantic-correlated tasks, we notice that in RMDE the relative changes in the least significant directions tend to be disproportionately large—primarily due to their near-zero base magnitudes—resulting in unstable and uninformative direction selection. Thus, we obtain directions with top $r$ values in $d$ as task-aware directions:
\begin{align}
\label{selection}
    \mathcal{I} = \mathrm{Top}\text{-}r(\mathbf{d}) \subset \{1,2, \dots, K\}, 
\end{align}
where $\mathcal{I}$ denotes the indices of the top-r singular directions with the highest projection magnitudes. Having the selected rank $r$ and direction list $\mathcal{I}$, low-rank tuning parameters can be intialized by
\begin{align}
\label{initialize}
    A &= \sqrt{\Sigma[\mathcal{I}]}V^T[\mathcal{I},:], \\
    B &= U[:,\mathcal{I}]\sqrt{\Sigma[\mathcal{I}]},
\end{align}
and the pretrained weight at the start of tuning becomes
\begin{align}
\label{pretrained_weight}
    W_0 &= W-U[:,\mathcal{I}] \Sigma[\mathcal{I}] V^T[\mathcal{I},:].
\end{align}

$[\mathcal{I}]$ denotes selecting the components from the indices $\mathcal{I}$. With the proper selection and initialization, the low-rank tuning can be more efficient in training $BA$ to adapt to the RMDE task.

Meanwhile, it is also crucial to preserve the generalization and robustness of the pretrained VFM during tuning. Given that the stable rank $Rank_{\text{st}}$ objectively reflects the concentration of energy within a weight matrix, we leverage it to identify the leading singular components to be preserved. These directions serve as structural anchors, preventing the adapted model from drifting away from the pretrained prior. This inspires us to design a component-preservation regularization for maintaining:
\begin{align}
\label{regularization}
    L_{\rm{reg}} = \frac{1}{N}\sum_{j=1}^{N}\sum_{i=1}^{Rank_{\rm{st}}}|\sigma_i u_i^T (BA) v_i|, \: \mathrm{for} \: i \le Rank_{\rm{st}} \: \& \: i \notin \mathcal{I},
\end{align}
where $N$ represents the number of layers. This regularization guarantees to protect the pretrained components, while not preventing the adaptation of those task-aware directions. Overall, the STM strategy balances task-specific learning and generalization preservation through entropy-rank and stable-rank-guided low-rank adaptation.

\subsection{Training Pipelines}
\label{sec:Training_Pipelines}
Our training pipeline is designed for both self-supervised learning and supervised learning in RMDE. In self-supervised RMDE, we follow the common practice in previous approaches \cite{gasperini_morbitzer2023md4all, godard2019digging} to design the objectives with Eq. \ref{equ:pe_loss} and a smooth loss as follows:
\begin{align}
\label{ssl_loss}
    L_{\rm{ssl}} = L_{\rm{pe}} + L_{\rm{smooth}}+L_{\rm{reg}}.
\end{align}

Considering the extreme sparsity of LiDAR data in the real-world dataset \cite{nuscenes2019, robotcar}, which is also utilized and mentioned by md4all \cite{gasperini_morbitzer2023md4all}, we add auxiliary pseudo labels generated by DepthAnything V2 \cite{depth_anything_v2}, to provide a dense supervisory signal.
The loss with LiDAR data $L_{\rm{gt}}$ is chosen as an absolute relative loss similar to \cite{gasperini_morbitzer2023md4all}, and the dense supervision from pseudo labels $L_{\rm{pseudo}}$ is chosen as a normalized L1 loss. Our supervised RMDE objective is designed as
\begin{align}
\label{ssl_loss}
    L_{\rm{sl}} = \alpha_1 L_{\rm{gt}}+ \alpha_2 L_{\rm{pseudo}} + L_{\rm{smooth}}+L_{\rm{reg}},
\end{align}
where we balance $\alpha_1 : \alpha_2=2:1$ to rely more on ground truth data. Noted that all the training is conducted with \textbf{a few daytime data only}, and generalizes to unseen adverse conditions, which is different from \cite{depth_anything_v2, depthanything} that learn from a huge amount of data with depth labels or pseudo labels.

\section{Experiments}
\label{experiments}


\subsection{Implementation Details}
\label{implementation}
In our experiments, by default, we use DINOv2-L \cite{oquab2023dinov2} as the frozen VFM, and DPT \cite{DPT2021} as the decoder. In the training, the learning rates for the backbone, decoder, and low-rank tuning are set to 5e-6, 2e-5, and 2e-5. All experiments are trained for $20$ epochs, except for full fine-tuning under supervised learning, we follow md4all \cite{gasperini_morbitzer2023md4all} to report results using the last checkpoint before the ripple artifacts caused by sparse ground truth supervision become evident.

\subsection{Dataset and Evaluation Protocol}

We evaluate our method on four real-world outdoor datasets under diverse weather and illumination conditions. Specifically, \textbf{NuScenes} (NS)~\cite{nuscenes2019} provides 15{,}129 daytime images for training and 6{,}019 testing images (including 4{,}449 from daytime, 1{,}088 from rain, and 602 from nighttime). \textbf{RobotCar} (RC)~\cite{robotcar} contains 17{,}790 daytime training images and 1{,}411 test images (702 daytime, 709 nighttime). \textbf{DrivingStereo} (DS)~\cite{drivingstereo} offers 500 rainy and 500 foggy images for zero-shot evaluation. \textbf{CADC}~\cite{cadc} contains 510 real snow images for zero-shot testing. All images are input to the models with a resolution of $574 \times 322$.

Our goal is to enable vision foundation models (VFMs) to generalize from normal daytime data to arbitrary unseen adverse conditions, without the need for synthetic adverse data. To this end, we train our model only on daytime images and evaluate on all unseen domains. For self-supervised experiments, we follow prior practices \cite{gasperini_morbitzer2023md4all, diffusionadverse} and incorporate a weak velocity loss to stabilize monocular training. The evaluation protocol follows MiDaS~\cite{birkl2023midas}, aligning all predictions to the ground truth using scale and shift before computing depth errors. We use an evaluation range of $0\text{--}80\,\mathrm{m}$ for all datasets except RobotCar, which uses $0\text{--}50\,\mathrm{m}$ following prior convention.

In the self-supervised setting, we compare with four categories of baselines: (1) existing self-supervised robust depth estimation methods \cite{gasperini_morbitzer2023md4all, Li_2023_ICCV, wang2021RNW, diffusionadverse}; (2) PEFT methods \cite{Rein, hu2022lora, yun2024soma} under self-supervised monocular depth frameworks; (3) full fine-tuning and freezing of VFMs on daytime data only; and (4) full fine-tuning of VFMs using synthetic data from \cite{gasperini_morbitzer2023md4all}. In the supervised setting, we compare against: (1) supervised RMDE approaches md4all~\cite{gasperini_morbitzer2023md4all}; (2) PEFT-based supervised tuning methods; (3) full fine-tuning of VFMs on labeled data; and (4) a state-of-the-art MDE model, Depth Anything v2~\cite{depth_anything_v2}. All methods are evaluated under the same protocol for fair comparison, with the normally used metrics: absolute relative error (AbsRel), square relative error (SqRel), root mean square error (RMSE), and threshold accuracy ($\delta_1)$.

\begin{table*}[tb] 
\renewcommand{\arraystretch}{1.2}
\caption{Self-supervised results. $^{\dag}$ indicates the utility of either synthetic adverse data or real-world adverse data that violates the photometric consistency. * denotes the number of trainable parameters inside the backbones. The first and second ranked performances are indicated by \textbf{bold} and \underline{underline}. }
\centering
\scriptsize
\setlength{\tabcolsep}{1pt}  
\begin{tabular}{l|c|c|cccc|cccc|cccc} 
\toprule 
\cline{1-15}\
\multirow{2}{*}{Method} &\multirow{2}{*}{Backbone} &\multicolumn{1}{c|}{Trainable}& \multicolumn{4}{c|}{\textbf{nuScenes-day}} & \multicolumn{4}{c|}{\textbf{nuScenes-night}} & \multicolumn{4}{c}{\textbf{nuScenes-rain}}\\
     & &Params* & AbsRel$\downarrow$ & SqRel$\downarrow$ &  RMSE$\downarrow$ & $ \delta_1\uparrow $ & AbsRel$\downarrow$ & SqRel$\downarrow$ &  RMSE$\downarrow$ & $ \delta_1\uparrow $ & AbsRel$\downarrow$ & SqRel$\downarrow$ &  RMSE$\downarrow$ & $ \delta_1\uparrow $    \\  
\cline{1-15}
Monodepth2 \cite{godard2019digging} & ResNet & 11.7M &13.33 &1.820 &6.459 &85.88 &24.19 &2.776 &10.922 &58.17 & 15.72 &2.273 &7.453 & 79.49 \\
RNW$^{\textbf{\dag}}$ \cite{wang2021RNW}  & ResNet & 11.7M &28.72 &3.433  &9.185 &56.21 &33.33 &4.066 &10.098 &43.72 &29.52 &3.796 &9.341 &57.21 \\
 robust-depth$^{\textbf{\dag}}$ \cite{Saunders_2023_ICCV} & ResNet & 11.7M &14.36 &1.862 &6.802 &82.95 &21.01 &2.691 &8.673 &69.58 &14.58 &1.891 &7.371 &80.21 \\
 md4all$^{\textbf{\dag}}$ \cite{gasperini_morbitzer2023md4all} & ResNet & 11.7M &13.66 &1.752 &6.452 &84.61 &19.21 & 2.386 &8.507 &71.07 &14.14 &1.829 &7.228 &80.98 \\
 DM-MDE$^{\textbf{\dag}}$ \cite{diffusionadverse} & ResNet & 11.7M &12.80 &- &6.449 &84.03 &19.10 &- &8.433 &71.14 &13.90 &- &7.129&81.36 \\
 \cline{1-15}
FFT & DINOv2 & 304.2M &11.87 &1.518 &6.117 &88.39 &18.73 &2.459 &8.242 &74.26 &13.69 &1.709 &6.725 & 84.03 \\
FFT+md4all$^{\textbf{\dag}}$ & DINOv2 & 304.2M &11.55 &1.412 &6.060 &88.54 &18.69 &2.593 &8.391 &73.99 &\underline{12.80} &1.675 &6.639 &\underline{84.98} \\
\cline{1-15}
Freeze & DINOv2 & 0.0M &12.00 &1.502 &6.086 &87.49 &\underline{17.22} &\underline{1.883} &8.051 &72.19 &13.37 &1.668 &6.580 &84.27 \\
Rein \cite{Rein} & DINOv2 & 5.0M &12.09 &1.475 &6.167 &87.55 &17.23 &1.896 &\underline{7.711} &73.52 &13.66 &1.707 & 6.673 & 84.48 \\
LoRA \cite{hu2022lora} & DINOv2 & 7.0M &11.40 &\textbf{1.247} &\underline{5.875} &88.32 &17.57 &1.978 &7.730 &74.40 &13.26 &\underline{1.572} &\underline{6.524} &84.61 \\
SoMA \cite{yun2024soma} & DINOv2 & 5.3M &\underline{11.34} &1.422 &6.018 &\underline{88.91} &17.88 &1.960 &7.732 &\underline{74.53} &13.05 & 1.705 & 6.545 & 84.77 \\
\cline{1-15}
\textbf{Ours} & DINOv2 & 8.7M &\textbf{10.78} &\underline{1.256} &\textbf{5.797} &\textbf{89.18} & \textbf{16.75} & \textbf{1.772} &\textbf{7.465} &\textbf{75.23} &\textbf{12.40} &\textbf{1.532} &\textbf{6.404} &\textbf{85.40} \\
\cline{1-15}
\bottomrule 
\end{tabular}
\label{tab:nuscenes}
\vspace{-0.2cm}
\end{table*}

\begin{table*}[t] 
\renewcommand{\arraystretch}{1.2}
\caption{Zero-shot evaluations for self-supervised methods on multiple datasets of diverse conditions. $\textbf{*}$ indicates the results on RobotCar dataset are based on models trained on the corresponding training data, while the others are conducted in a zero-shot setting.}
\centering
\scriptsize
\setlength{\tabcolsep}{.6pt}  
\begin{tabular}{l|ccc|ccc|ccc|ccc|ccc} 
\toprule 
\cline{1-16}\
\multirow{2}{*}{Method} & \multicolumn{3}{c|}{\textbf{RC-day}} & \multicolumn{3}{c|}{\textbf{RC-night}} & \multicolumn{3}{c|}{\textbf{DS-rain}} & \multicolumn{3}{c|}{\textbf{DS-fog}} & \multicolumn{3}{c}{\textbf{CADC-Snow}}\\
     & AbsRel$\downarrow$  &  RMSE$\downarrow$ & $ \delta_1\uparrow $ & AbsRel$\downarrow$  &  RMSE$\downarrow$ & $ \delta_1\uparrow $ & AbsRel$\downarrow$  &  RMSE$\downarrow$ & $ \delta_1\uparrow $ &
     AbsRel$\downarrow$  &  RMSE$\downarrow$ & $ \delta_1\uparrow $
     &
     AbsRel$\downarrow$  &  RMSE$\downarrow$ & $ \delta_1\uparrow $\\  
\cline{1-16}
    robust-depth$\textbf{*}$ \cite{Saunders_2023_ICCV} &12.25 &3.302 &85.84 &13.33 &3.756 &85.11 &26.22 &11.657 &59.58 &13.66 &6.876 &84.01 &31.47  &13.65 &43.37\\
 md4all$\textbf{*}$ \cite{gasperini_morbitzer2023md4all} &11.28 &3.206 &87.13 &12.19 &3.604 & 84.86 &18.22 &8.465 &70.35 &12.42 &6.269 &86.16 & 29.65 &12.85 &48.59 \\
  DM-MDE$\textbf{*}$ \cite{diffusionadverse} &11.90 &3.287 &87.17 &12.90 &3.661 & 83.68 &- &- &- &- &- &- & - &- &- \\
\cline{1-16}
FFT &10.36 &3.099 &89.20 &12.31 &3.667 &87.53 &\underline{10.57} &\textbf{5.262} &90.76 &8.69 &4.839 & \underline{94.02}
&\underline{25.76}  &10.44 &65.93\\
FFT+md4all  &10.08 &3.118 &89.15 &12.70 &3.973 &84.37 &11.08 &\underline{5.394} &89.15 &8.71 &\underline{4.662} &93.99 &26.24 &10.49 &65.10 \\
\cline{1-16}
Freeze &9.54 &\textbf{2.906} &90.03 &11.74 &3.687 &85.64 &10.63 &5.476 &\underline{90.91} &\textbf{8.50} &4.740 &93.92 &27.23 &\underline{10.37} &61.98 \\
Rein \cite{Rein} &9.62 &2.986 &90.14  &12.13 &3.766 &85.87 &11.42 &5.739 &88.98 &\underline{8.55} &4.987 &93.69 &26.80 &10.81 &64.51\\
LoRA \cite{hu2022lora}  &9.52 &3.018 &90.31 &11.53 &3.458 &88.07 &11.17 &5.545 &90.20 &8.89 &4.811 &93.86 &26.07 &10.42 &65.11 \\
SoMA \cite{yun2024soma}  &\textbf{9.39} &2.920 &\textbf{90.73} &\underline{11.43} &\underline{3.401} &\underline{88.28} &11.60 &6.079 &88.90 &9.67 &5.342 &92.94 &26.10 &10.62 &\underline{66.11} \\
\cline{1-16}
\textbf{Ours} &\underline{9.41} &\underline{2.917} &\underline{90.57} &\textbf{11.12} & \textbf{3.393} & \textbf{88.81} &\textbf{10.49} &5.444 &\textbf{91.45} &8.62 &\textbf{4.659} &\textbf{94.42} & \textbf{24.36}  &\textbf{10.08} &\textbf{67.71}\\
\cline{1-16}
\bottomrule 
\end{tabular}
\label{tab:zero-shot}
\end{table*}

\begin{table*}[t] 
\renewcommand{\arraystretch}{1.2}
\caption{Quantitative results under supervised setting. Noted that Depth Anything v2 \cite{depth_anything_v2} is trained with more than $ 62\mathrm {M}$ of data, while our model only learns from $15\rm{K}$ daytime images.}
\centering
\scriptsize
\setlength{\tabcolsep}{.4pt}  
\begin{tabular}{l|ccc|ccc|ccc|ccc|ccc|ccc} 
\toprule 
\cline{1-19}\
\multirow{2}{*}{Method} & \multicolumn{3}{c|}{\textbf{NS-night}} & \multicolumn{3}{c|}{\textbf{NS-rain}} & \multicolumn{3}{c|}{\textbf{RC-night}} & \multicolumn{3}{c|}{\textbf{DS-rain}} & \multicolumn{3}{c|}{\textbf{DS-fog}} & \multicolumn{3}{c}{\textbf{CADC-snow}}\\
     & AbsRel  &  RMSE & $ \delta_1 $ & AbsRel  &  RMSE & $ \delta_1 $ & AbsRel  &  RMSE & $ \delta_1 $ &
     AbsRel  &  RMSE & $ \delta_1 $
     &
     AbsRel &  RMSE & $ \delta_1 $ & AbsRel &  RMSE & $ \delta_1 $\\  
\cline{1-19}
 md4all \cite{gasperini_morbitzer2023md4all} &18.21 &6.372 &75.33 &15.62 &5.903 & 82.82 &- &- &- &- &- &- & - &- &- &- &- &- \\
FFT &\underline{14.88} &\textbf{7.214} &\underline{78.37} &\underline{8.55} &\underline{5.290} &\underline{90.13} &\underline{8.74} &2.893 &90.53 & 8.69 &4.889 & 93.15 &6.18 &4.096 &95.31 &22.61 &10.44 &66.21 \\
Freeze &17.57 &8.241 &73.20 &9.60 &5.520 &87.91 &8.99 &2.908 &89.71 &9.30 &5.008 &92.07 &6.76 &4.353 &94.38  &\underline{21.90} & \textbf{9.973} &\underline{68.13}  \\
Rein \cite{Rein} &18.05 &8.424 &72.53 &9.75 &5.615 &87.51 &9.25 &\textbf{2.891} &89.56 &9.22 &5.009 &92.33 &6.65 &4.367 &94.47 &22.34 &10.25 &66.22 \\
LoRA \cite{hu2022lora}  &15.38 &7.379 &77.74 &8.66 &5.341 &89.89 &8.87 &2.914 &90.33 &\underline{8.07} &\underline{4.459} &\underline{94.32} &\underline{5.99} &\underline{3.855} &\underline{95.97} &21.97 &\underline{10.07} &66.33 \\
\cline{1-19}
DA v2 \cite{depth_anything_v2} &17.64 &8.424 &74.87 &12.19 &7.217 &84.51 &\textbf{8.60} &3.150 &\textbf{91.66} &9.04 &5.030 &92.89 &6.88 &4.552 &95.43 &22.53 &10.78 &\textbf{72.97} \\
\cline{1-19}
\textbf{Ours} &\textbf{14.82}  &\underline{7.258} &\textbf{78.57} &\textbf{8.45} &\textbf{5.218} & \textbf{90.24} &8.85 &\underline{2.893} &\underline{90.63} &\textbf{7.81} &\textbf{4.349} &\textbf{94.90} &\textbf{5.71} &\textbf{3.729} &\textbf{96.30} &\textbf{21.73} &10.10 &66.37 \\
\cline{1-19}
\bottomrule 
\end{tabular}
\label{tab:supervised}
\vspace{-0.2cm}
\end{table*}

\subsection{Experimental Results}
Tab.~\ref{tab:nuscenes} reports quantitative results under the self-supervised setting on the nuScenes~\cite{nuscenes2019} dataset. Our method achieves the best overall performance across unseen adverse domains (nighttime and rain). Compared to prior PEFT methods, our approach reduces the average AbsRel of 5.5\%, and RMSE of 3.1\%, while using similar trainable parameters. Against synthetic-data-based robust depth baselines, we achieve significantly better robustness. Noted that full fine-tuning leads to noticeable degradation of the robust priors encoded in the pretrained VFM. Moreover, introducing synthetic adverse data tends to interfere with the knowledge learned from large-scale natural image distributions, ultimately harming the VFM's generalization.

Tab.~\ref{tab:zero-shot} presents zero-shot results on five unseen domains, learning from only nuScenes-daytime. Our method achieves the best overall generalization, outperforming all prior methods across average AbsRel, RMSE, and $\delta_1$. Compared to the latest PEFT method, SoMA~\cite{yun2024soma}, our approach improves average AbsRel and RMSE by 5.1\% and 5.7\%, respectively, and surpasses full fine-tuning (FFT) by 4.4\% and 2.1\%. These validate the robustness of our STM strategy under diverse distribution shifts.

As shown in Table~\ref{tab:supervised}, our method, trained on only 15K daytime images, consistently outperforms all baselines, including PEFT, FFT, and Depth Anything V2 \cite{depth_anything_v2}. It achieves an average improvement of 7.4\% AbsRel and 6.1\% RMSE, demonstrating superior generalization under limited supervision.

\begin{figure*}[!tb]
\centering
\includegraphics[width=.98\linewidth, height=.32\linewidth]{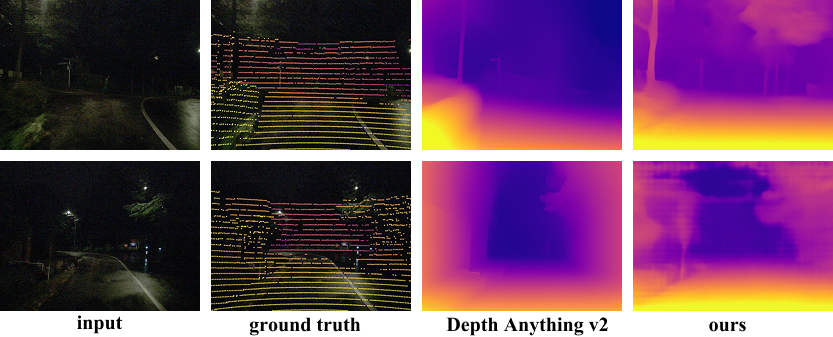}
\caption{Qualitative Results under extreme conditions.}
\label{fig:qualitative}
\vspace{-0.3cm}
\end{figure*}

\subsection{Ablations}
 To validate the effectiveness of each proposed component, we conduct an ablation study with controlled variants. Specifically, $Rank_{\rm{en}}$ denotes dynamic assignment of rank numbers per weight matrix, $Task_{\rm{dir}}$ refers to selecting task-aware directions for initialization, and $Rank_{\rm{st}}$ means applying regularization on important singular directions. The results are summarized in Table~\ref{table:ablation}.

\begin{table*}[htb] 
\vspace{-0.2cm}
\renewcommand{\arraystretch}{1.1}
\centering
\small
\resizebox{.92\linewidth}{!}{
\begin{tabular}{ccc|cccc|cccc} 
\toprule
\cline{1-11}
\multicolumn{3}{c|}{Modules }  & \multicolumn{4}{c|}{nuScenes-night} & \multicolumn{4}{c}{nuScenes-rain}\\
\textbf{$Rank_{\rm{en}}$} & $Task_{\rm{dir}}$ & $Rank_{\rm{st}}$ & AbsRel$\downarrow$ & SqRel$\downarrow$ & RMSE$\downarrow$ & $ \delta_1\uparrow $ & AbsRel$\downarrow$ & SqRel$\downarrow$ & RMSE$\downarrow$ & $ \delta_1\uparrow $  \\  
\cline{1-11}
\checkmark &&& 17.21 & 2.087 & 7.693 & 74.76 & 12.93 & 1.633 & 6.485 & 84.86 \\
\checkmark & \checkmark && 16.96 & 1.942 & 7.604 & 75.06 & 12.58 & 1.596 & 6.438 & 85.12 \\
\checkmark & \checkmark & \checkmark & \textbf{16.75} & \textbf{1.772} & \textbf{7.465} & \textbf{75.23} & \textbf{12.40} & \textbf{1.532} & \textbf{6.404} & \textbf{85.40} \\
\hline
\bottomrule 
\end{tabular}}
\vspace{-0.1cm}
\caption{Ablations on each module mentioned in the pipeline. The checkmark represents the utility of the corresponding module.}
\label{table:ablation}
\vspace{-0.3cm}
\end{table*}

Overall, these modules are designed to jointly enhance generalization and stability: $Rank_{\rm{en}}$ adapts model capacity to each weight matrix’s structure, $Task_{\rm{dir}}$ ensures alignment between initialization and downstream tasks, and $Rank_{\rm{st}}$ preserves essential representational subspaces during adaptation.

\section{Conclusion}
\label{conclusion}
In this work, we propose ER-LoRA, a novel parameter-efficient fine-tuning (PEFT) pipeline for robust monocular depth estimation under diverse weather conditions. By analyzing the singular value spectrum of vision foundation models (VFMs), we introduce a Selecting–Tuning–Maintaining (STM) strategy that leverages effective rank to guide task-relevant initialization, discriminative tuning, and structural regularization. This design enables ER-LoRA to adapt to unseen domains without relying on synthetic data, while preserving the strong generalization ability of pretrained VFMs. Experiments on diverse real-world datasets under varying conditions validate the effectiveness of our method in both self-supervised and supervised settings, outperforming a wide range of existing approaches across different methodological branches.

\begin{ack}
Use unnumbered first level headings for the acknowledgments. All acknowledgments
go at the end of the paper before the list of references. Moreover, you are required to declare
funding (financial activities supporting the submitted work) and competing interests (related financial activities outside the submitted work).
More information about this disclosure can be found at: \url{https://neurips.cc/Conferences/2025/PaperInformation/FundingDisclosure}.

Do {\bf not} include this section in the anonymized submission, only in the final paper. You can use the \texttt{ack} environment provided in the style file to automatically hide this section in the anonymized submission.
\end{ack}







{\small
\bibliographystyle{plain}
\bibliography{main}

\begin{thebibliography}{10}

\bibitem{domain_human_pose}
Yihao Ai, Yifei Qi, Bo~Wang, Yu~Cheng, Xinchao Wang, and Robby~T Tan.
\newblock Domain-adaptive 2d human pose estimation via dual teachers in extremely low-light conditions.
\newblock In {\em European Conference on Computer Vision}, pages 221--239. Springer, 2024.

\bibitem{zoedepth}
Shariq~Farooq Bhat, Reiner Birkl, Diana Wofk, Peter Wonka, and Matthias Müller.
\newblock Zoedepth: Zero-shot transfer by combining relative and metric depth, 2023.

\bibitem{bi2024fada}
Qi~Bi, Jingjun Yi, Hao Zheng, Haolan Zhan, Yawen Huang, Wei Ji, Yuexiang Li, and Yefeng Zheng.
\newblock Learning frequency-adapted vision foundation model for domain generalized semantic segmentation.
\newblock In {\em Advances in Neural Information Processing Systems (NeurIPS)}, volume~37, pages 94047--94072, 2024.

\bibitem{bian2021ijcv_scdepth}
Jia-Wang Bian, Huangying Zhan, Naiyan Wang, Zhichao Li, Le~Zhang, Chunhua Shen, Ming-Ming Cheng, and Ian Reid.
\newblock Unsupervised scale-consistent depth learning from video.
\newblock {\em International Journal of Computer Vision (IJCV)}, 2021.

\bibitem{birkl2023midas}
Reiner Birkl, Diana Wofk, and Matthias M{\"u}ller.
\newblock Midas v3.1 -- a model zoo for robust monocular relative depth estimation.
\newblock {\em arXiv preprint arXiv:2307.14460}, 2023.

\bibitem{nuscenes2019}
Holger Caesar, Varun Bankiti, Alex~H. Lang, Sourabh Vora, Venice~Erin Liong, Qiang Xu, Anush Krishnan, Yu~Pan, Giancarlo Baldan, and Oscar Beijbom.
\newblock nuscenes: A multimodal dataset for autonomous driving.
\newblock {\em arXiv preprint arXiv:1903.11027}, 2019.

\bibitem{ssnerf}
Xiao Cao, Beibei Lin, Bo~Wang, Zhiyong Huang, and Robby~T. Tan.
\newblock Ssnerf: Sparse view semi-supervised neural radiance fields with augmentation.
\newblock {\em arXiv preprint arXiv:2408.09144}, 2024.

\bibitem{3dswapping}
Xiao Cao, Beibei Lin, Bo~Wang, Zhiyong Huang, and Robby~T. Tan.
\newblock 3dswapping: Texture swapping for 3d object from single reference image.
\newblock {\em arXiv preprint arXiv:2503.18853}, 2025.

\bibitem{languageguidance}
Agneet Chatterjee, Tejas Gokhale, Chitta Baral, and Yezhou Yang.
\newblock On the robustness of language guidance for low-level vision tasks: Findings from depth estimation.
\newblock In {\em Proceedings of the IEEE/CVF Conference on Computer Vision and Pattern Recognition (CVPR)}, pages 2794--2803, June 2024.

\bibitem{chen2022adaptformer}
Shoufa Chen, Chongjian Ge, Zhan Tong, Jiangliu Wang, Yibing Song, Jue Wang, and Ping Luo.
\newblock Adaptformer: Adapting vision transformers for scalable visual recognition.
\newblock {\em arXiv preprint arXiv:2205.13535}, 2022.

\bibitem{Eckart_Young_1936}
Carl Eckart and Gale Young.
\newblock The approximation of one matrix by another of lower rank.
\newblock {\em Psychometrika}, 1(3):211–218, 1936.

\bibitem{Eigen2014}
David Eigen, Christian Puhrsch, and Rob Fergus.
\newblock Depth map prediction from a single image using a multi-scale deep network.
\newblock In {\em Proceedings of the 28th International Conference on Neural Information Processing Systems - Volume 2}, NIPS'14, page 2366–2374, Cambridge, MA, USA, 2014. MIT Press.

\bibitem{fan2025makeloragreatagain}
Chenghao Fan, Zhenyi Lu, Sichen Liu, Xiaoye Qu, Wei Wei, Chengfeng Gu, and Yu~Cheng.
\newblock Make lora great again: Boosting lora with adaptive singular values and mixture-of-experts optimization alignment, 2025.

\bibitem{DCLdepth}
Junkai Fan, Kun Wang, Zhiqiang Yan, Xiang Chen, Shangbing Gao, Jun Li, and Jian Yang.
\newblock Depth-centric dehazing and depth-estimation from real-world hazy driving video.
\newblock In {\em Proceedings of the AAAI Conference on Artificial Intelligence}, pages xxxxx--xxxxx, 2025.

\bibitem{EVA}
Yuxin Fang, Wen Wang, Binhui Xie, Quan Sun, Ledell Wu, Xinggang Wang, Tiejun Huang, Xinlong Wang, and Yue Cao.
\newblock Eva: Exploring the limits of masked visual representation learning at scale.
\newblock {\em arXiv preprint arXiv:2211.07636}, 2022.

\bibitem{fu2024geowizard}
Xiao Fu, Wei Yin, Mu~Hu, Kaixuan Wang, Yuexin Ma, Ping Tan, Shaojie Shen, Dahua Lin, and Xiaoxiao Long.
\newblock Geowizard: Unleashing the diffusion priors for 3d geometry estimation from a single image.
\newblock In {\em European Conference on Computer Vision}, pages 241--258. Springer, 2024.

\bibitem{gasperini_morbitzer2023md4all}
Stefano Gasperini, Nils Morbitzer, HyunJun Jung, Nassir Navab, and Federico Tombari.
\newblock Robust monocular depth estimation under challenging conditions.
\newblock In {\em Proceedings of the IEEE/CVF International Conference on Computer Vision}, pages 8177--8186, 2023.

\bibitem{godard2019digging}
Cl{\'e}ment Godard, Oisin Mac~Aodha, Michael Firman, and Gabriel~J Brostow.
\newblock Digging into self-supervised monocular depth estimation.
\newblock In {\em Proceedings of the IEEE/CVF international conference on computer vision}, pages 3828--3838, 2019.

\bibitem{2017left-right-consistency}
Clément Godard, Oisin~Mac Aodha, and Gabriel~J. Brostow.
\newblock Unsupervised monocular depth estimation with left-right consistency.
\newblock In {\em 2017 IEEE Conference on Computer Vision and Pattern Recognition (CVPR)}, pages 6602--6611, 2017.

\bibitem{gui2024depthfm}
Ming Gui, Johannes Schusterbauer, Ulrich Prestel, Pingchuan Ma, Dmytro Kotovenko, Olga Grebenkova, Stefan~Andreas Baumann, Vincent~Tao Hu, and Björn Ommer.
\newblock Depthfm: Fast monocular depth estimation with flow matching, 2024.

\bibitem{MAE}
Kaiming He, Xinlei Chen, Saining Xie, Yanghao Li, Piotr Doll{\'a}r, and Ross Girshick.
\newblock Masked autoencoders are scalable vision learners.
\newblock {\em arXiv:2111.06377}, 2021.

\bibitem{he2015delvingdeeprectifierssurpassing}
Kaiming He, Xiangyu Zhang, Shaoqing Ren, and Jian Sun.
\newblock Delving deep into rectifiers: Surpassing human-level performance on imagenet classification, 2015.

\bibitem{hu2022lora}
Edward~J Hu, Yelong Shen, Phillip Wallis, Zeyuan Allen-Zhu, Yuanzhi Li, Shean Wang, Lu~Wang, and Weizhu Chen.
\newblock Lo{RA}: Low-rank adaptation of large language models.
\newblock In {\em International Conference on Learning Representations}, 2022.

\bibitem{hu2024metric3d}
Mu~Hu, Wei Yin, Chi Zhang, Zhipeng Cai, Xiaoxiao Long, Hao Chen, Kaixuan Wang, Gang Yu, Chunhua Shen, and Shaojie Shen.
\newblock Metric3d v2: A versatile monocular geometric foundation model for zero-shot metric depth and surface normal estimation.
\newblock {\em IEEE Transactions on Pattern Analysis and Machine Intelligence}, 2024.

\bibitem{jia2022visualprompttuning}
Menglin Jia, Luming Tang, Bor-Chun Chen, Claire Cardie, Serge Belongie, Bharath Hariharan, and Ser-Nam Lim.
\newblock Visual prompt tuning, 2022.

\bibitem{marigold}
Bingxin Ke, Anton Obukhov, Shengyu Huang, Nando Metzger, Rodrigo~Caye Daudt, and Konrad Schindler.
\newblock Repurposing diffusion-based image generators for monocular depth estimation.
\newblock In {\em Proceedings of the IEEE/CVF Conference on Computer Vision and Pattern Recognition (CVPR)}, 2024.

\bibitem{depth_resnet2016}
Iro Laina, Christian Rupprecht, Vasileios Belagiannis, Federico Tombari, and Nassir Navab.
\newblock Deeper depth prediction with fully convolutional residual networks.
\newblock In {\em 2016 Fourth International Conference on 3D Vision (3DV)}, pages 239--248, 2016.

\bibitem{lester2021powerscaleparameterefficientprompt}
Brian Lester, Rami Al-Rfou, and Noah Constant.
\newblock The power of scale for parameter-efficient prompt tuning, 2021.

\bibitem{Li_2023_ICCV}
Ming Li, Xiangyu Xu, Hehe Fan, Pan Zhou, Jun Liu, Jia-Wei Liu, Jiahe Li, Jussi Keppo, Mike~Zheng Shou, and Shuicheng Yan.
\newblock Stprivacy: Spatio-temporal privacy-preserving action recognition.
\newblock In {\em Proceedings of the IEEE/CVF International Conference on Computer Vision (ICCV)}, pages 5106--5115, October 2023.

\bibitem{li2022depthformer}
Zhenyu Li, Zehui Chen, Xianming Liu, and Junjun Jiang.
\newblock Depthformer: Exploiting long-range correlation and local information for accurate monocular depth estimation.
\newblock {\em arXiv preprint arXiv:2203.14211}, 2022.

\bibitem{liu2021ADIDS}
Lina Liu, Xibin Song, Mengmeng Wang, Yong Liu, and Liangjun Zhang.
\newblock Self-supervised monocular depth estimation for all day images using domain separation.
\newblock In {\em Proceedings of the IEEE/CVF International Conference on Computer Vision}, pages 12737--12746, 2021.

\bibitem{liu2024dora}
Shih-Yang Liu, Chien-Yi Wang, Hongxu Yin, Pavlo Molchanov, Yu-Chiang~Frank Wang, Kwang-Ting Cheng, and Min-Hung Chen.
\newblock Dora: Weight-decomposed low-rank adaptation.
\newblock {\em arXiv preprint arXiv:2402.09353}, 2024.

\bibitem{robotcar}
W.~Maddern, G.~Pascoe, C.~Linegar, and P.~Newman.
\newblock 1 year, 1000 km: The oxford robotcar dataset.
\newblock {\em International Journal of Robotics Research}, page 0278364916679498, 2016.

\bibitem{meng2024pissa}
Fanxu Meng, Zhaohui Wang, and Muhan Zhang.
\newblock Pissa: Principal singular values and singular vectors adaptation of large language models.
\newblock {\em arXiv preprint arXiv:2404.02948}, 2024.

\bibitem{oquab2023dinov2}
Maxime Oquab, Timothée Darcet, Theo Moutakanni, Huy~V. Vo, Marc Szafraniec, Vasil Khalidov, Pierre Fernandez, Daniel Haziza, Francisco Massa, Alaaeldin El-Nouby, Russell Howes, Po-Yao Huang, Hu~Xu, Vasu Sharma, Shang-Wen Li, Wojciech Galuba, Mike Rabbat, Mido Assran, Nicolas Ballas, Gabriel Synnaeve, Ishan Misra, Herve Jegou, Julien Mairal, Patrick Labatut, Armand Joulin, and Piotr Bojanowski.
\newblock Dinov2: Learning robust visual features without supervision, 2023.

\bibitem{ecodepth}
Suraj Patni, Aradhye Agarwal, and Chetan Arora.
\newblock Ecodepth: Effective conditioning of diffusion models for monocular depth estimation.
\newblock In {\em Proceedings of the IEEE/CVF Conference on Computer Vision and Pattern Recognition (CVPR)}, pages 28285--28295, June 2024.

\bibitem{2021epcdepth}
Rui Peng, Ronggang Wang, Yawen Lai, Luyang Tang, and Yangang Cai.
\newblock Excavating the potential capacity of self-supervised monocular depth estimation.
\newblock In {\em Proceedings of the IEEE International Conference on Computer Vision (ICCV)}, 2021.

\bibitem{cadc}
Matthew Pitropov, Danson~Evan Garcia, Jason Rebello, Michael Smart, Carlos Wang, Krzysztof Czarnecki, and Steven~Lake Waslander.
\newblock Canadian adverse driving conditions dataset.
\newblock {\em The International Journal of Robotics Research}, 2021.

\bibitem{clip}
Alec Radford, Jong~Wook Kim, Chris Hallacy, Aditya Ramesh, Gabriel Goh, Sandhini Agarwal, Girish Sastry, Amanda Askell, Pamela Mishkin, Jack Clark, Gretchen Krueger, and Ilya Sutskever.
\newblock Learning transferable visual models from natural language supervision.
\newblock {\em CoRR}, abs/2103.00020, 2021.

\bibitem{DPT2021}
Ren\'{e} Ranftl, Alexey Bochkovskiy, and Vladlen Koltun.
\newblock Vision transformers for dense prediction.
\newblock {\em ArXiv preprint}, 2021.

\bibitem{stablerank1}
Benjamin Recht, Maryam Fazel, and Pablo~A. Parrilo.
\newblock Guaranteed minimum-rank solutions of linear matrix equations via nuclear norm minimization.
\newblock {\em SIAM Review}, 52(3):471–501, January 2010.

\bibitem{LDM}
Robin Rombach, Andreas Blattmann, Dominik Lorenz, Patrick Esser, and Bj\"orn Ommer.
\newblock High-resolution image synthesis with latent diffusion models.
\newblock In {\em Proceedings of the IEEE/CVF Conference on Computer Vision and Pattern Recognition (CVPR)}, pages 10684--10695, June 2022.

\bibitem{entropyrank}
Olivier Roy and Martin Vetterli.
\newblock The effective rank: A measure of effective dimensionality.
\newblock In {\em 2007 15th European Signal Processing Conference}, pages 606--610, 2007.

\bibitem{stablerank3}
Mark Rudelson and Roman Vershynin.
\newblock Sampling from large matrices: an approach through geometric functional analysis, 2006.

\bibitem{Saunders_2023_ICCV}
Kieran Saunders, George Vogiatzis, and Luis~J. Manso.
\newblock Self-supervised monocular depth estimation: Let's talk about the weather.
\newblock In {\em Proceedings of the IEEE/CVF International Conference on Computer Vision (ICCV)}, pages 8907--8917, October 2023.

\bibitem{schon2021mgnet}
Markus Sch{\"o}n, Michael Buchholz, and Klaus Dietmayer.
\newblock Mgnet: Monocular geometric scene understanding for autonomous driving.
\newblock In {\em Proceedings of the IEEE/CVF International Conference on Computer Vision}, pages 15804--15815, 2021.

\bibitem{shao2023IEBins}
Shuwei Shao, Zhongcai Pei, Xingming Wu, Zhong Liu, Weihai Chen, and Zhengguo Li.
\newblock Iebins: Iterative elastic bins for monocular depth estimation.
\newblock In {\em Advances in Neural Information Processing Systems (NeurIPS)}, 2023.

\bibitem{pvchat}
Yufei Shi, Weilong Yan, Gang Xu, Yumeng Li, Yucheng Chen, Zhenxi Li, Fei~Richard Yu, Ming Li, and Si~Yong Yeo.
\newblock Pvchat: Personalized video chat with one-shot learning.
\newblock {\em arXiv preprint arXiv:2503.17069}, 2025.

\bibitem{task_specific_direction}
Chongjie Si, Zhiyi Shi, Shifan Zhang, Xiaokang Yang, Hanspeter Pfister, and Wei Shen.
\newblock Task-specific directions: Definition, exploration, and utilization in parameter efficient fine-tuning, 2025.

\bibitem{spencer2020defeat}
Jaime Spencer, Richard Bowden, and Simon Hadfield.
\newblock Defeat-net: General monocular depth via simultaneous unsupervised representation learning.
\newblock In {\em Proceedings of the IEEE/CVF Conference on Computer Vision and Pattern Recognition}, pages 14402--14413, 2020.

\bibitem{sc_depthv3}
Libo Sun, Jia-Wang Bian, Huangying Zhan, Wei Yin, Ian Reid, and Chunhua Shen.
\newblock Sc-depthv3: Robust self-supervised monocular depth estimation for dynamic scenes.
\newblock {\em IEEE Transactions on Pattern Analysis and Machine Intelligence (TPAMI)}, 2023.

\bibitem{diffusionadverse}
Fabio Tosi, Pierluigi Zama~Ramirez, and Matteo Poggi.
\newblock Diffusion models for monocular depth estimation: Overcoming challenging conditions.
\newblock In {\em European Conference on Computer Vision (ECCV)}, 2024.

\bibitem{stablerank2}
Joel~A. Tropp.
\newblock An introduction to matrix concentration inequalities, 2015.

\bibitem{vankadari2020ADFA}
Madhu Vankadari, Sourav Garg, Anima Majumder, Swagat Kumar, and Ardhendu Behera.
\newblock Unsupervised monocular depth estimation for night-time images using adversarial domain feature adaptation.
\newblock In {\em Computer Vision--ECCV 2020: 16th European Conference, Glasgow, UK, August 23--28, 2020, Proceedings, Part XXVIII 16}, pages 443--459. Springer, 2020.

\bibitem{vankadari2023sun}
Madhu Vankadari, Stuart Golodetz, Sourav Garg, Sangyun Shin, Andrew Markham, and Niki Trigoni.
\newblock When the sun goes down: Repairing photometric losses for all-day depth estimation.
\newblock In {\em Conference on Robot Learning}, pages 1992--2003. PMLR, 2023.

\bibitem{miLoRA}
Hanqing Wang, Yixia Li, Shuo Wang, Guanhua Chen, and Yun Chen.
\newblock Milora: Harnessing minor singular components for parameter-efficient llm finetuning, 2024.

\bibitem{DAR}
Jinhong Wang, Jian Liu, Dongqi Tang, Weiqiang Wang, Wentong Li, Danny Chen, Jintai Chen, and Jian Wu.
\newblock Scalable autoregressive monocular depth estimation.
\newblock {\em arXiv preprint arXiv:2411.11361}, 2024.

\bibitem{wang2023weatherdepth}
Jiyuan Wang, Chunyu Lin, Lang Nie, Shujun Huang, Yao Zhao, Xing Pan, and Rui Ai.
\newblock Weatherdepth: Curriculum contrastive learning for self-supervised depth estimation under adverse weather conditions, 2023.

\bibitem{diffusion_contrast}
Jiyuan Wang, Lang Nie, Kang Liao, Shuwei Shao, and Yao Zhao.
\newblock Digging into contrastive learning for robust depth estimation with diffusion models.
\newblock In {\em ACM Int. Conf. Multimedia (ACMMM)}, pages 4129--4137, 10 2024.

\bibitem{wang2021RNW}
Kun Wang, Zhenyu Zhang, Zhiqiang Yan, Xiang Li, Baobei Xu, Jun Li, and Jian Yang.
\newblock Regularizing nighttime weirdness: Efficient self-supervised monocular depth estimation in the dark.
\newblock In {\em Proceedings of the IEEE/CVF International Conference on Computer Vision}, pages 16055--16064, 2021.

\bibitem{2023planedepth}
Ruoyu Wang, Zehao Yu, and Shenghua Gao.
\newblock Planedepth: Self-supervised depth estimation via orthogonal planes.
\newblock In {\em 2023 IEEE/CVF Conference on Computer Vision and Pattern Recognition (CVPR)}, pages 21425--21434, 2023.

\bibitem{wang2025monopcc}
Zhiwei Wang, Ying Zhou, Shiquan He, Ting Li, Fan Huang, Qiang Ding, Xinxia Feng, Mei Liu, and Qiang Li.
\newblock Monopcc: Photometric-invariant cycle constraint for monocular depth estimation of endoscopic images.
\newblock {\em Medical Image Analysis}, 102:103534, 2025.

\bibitem{manydepth}
Jamie Watson, Oisin Mac~Aodha, Victor Prisacariu, Gabriel Brostow, and Michael Firman.
\newblock The temporal opportunist: Self-supervised multi-frame monocular depth.
\newblock In {\em Proceedings of the IEEE/CVF Conference on Computer Vision and Pattern Recognition (CVPR)}, pages 1164--1174, June 2021.

\bibitem{Rein}
Zhixiang Wei, Lin Chen, Yi~Jin, Xiaoxiao Ma, Tianle Liu, Pengyang Ling, Ben Wang, Huaian Chen, and Jinjin Zheng.
\newblock Stronger fewer \& superior: Harnessing vision foundation models for domain generalized semantic segmentation.
\newblock In {\em Proceedings of the IEEE/CVF Conference on Computer Vision and Pattern Recognition (CVPR)}, pages 28619--28630, June 2024.

\bibitem{syn2real-depth}
Weilong Yan, Ming Li, Haipeng Li, Shuwei Shao, and Robby~T. Tan.
\newblock Synthetic-to-real self-supervised robust depth estimation via learning with motion and structure priors.
\newblock {\em arXiv preprint arXiv:2503.20211}, 2025.

\bibitem{deep-homo}
Weilong Yan, Robby~T. Tan, Bing Zeng, and Shuaicheng Liu.
\newblock Deep homography mixture for single image rolling shutter correction.
\newblock In {\em 2023 IEEE/CVF International Conference on Computer Vision (ICCV)}, pages 9834--9843, 2023.

\bibitem{transdepth}
Guanglei Yang, Hao Tang, Mingli Ding, Nicu Sebe, and Elisa Ricci.
\newblock Transformer-based attention networks for continuous pixel-wise prediction.
\newblock In {\em ICCV}, 2021.

\bibitem{drivingstereo}
Guorun Yang, Xiao Song, Chaoqin Huang, Zhidong Deng, Jianping Shi, and Bolei Zhou.
\newblock Drivingstereo: A large-scale dataset for stereo matching in autonomous driving scenarios.
\newblock In {\em IEEE Conference on Computer Vision and Pattern Recognition (CVPR)}, 2019.

\bibitem{depthanything}
Lihe Yang, Bingyi Kang, Zilong Huang, Xiaogang Xu, Jiashi Feng, and Hengshuang Zhao.
\newblock Depth anything: Unleashing the power of large-scale unlabeled data.
\newblock In {\em CVPR}, 2024.

\bibitem{depth_anything_v2}
Lihe Yang, Bingyi Kang, Zilong Huang, Zhen Zhao, Xiaogang Xu, Jiashi Feng, and Hengshuang Zhao.
\newblock Depth anything v2.
\newblock {\em arXiv:2406.09414}, 2024.

\bibitem{set}
Jingjun Yi, Qi~Bi, Hao Zheng, Haolan Zhan, Wei Ji, Yawen Huang, Yuexiang Li, and Yefeng Zheng.
\newblock Learning spectral-decomposed tokens for domain generalized semantic segmentation.
\newblock {\em arXiv preprint arXiv:2407.18568}, 2024.

\bibitem{yu2023udepth}
Boxiao Yu, Jiayi Wu, and Md~Jahidul Islam.
\newblock Udepth: Fast monocular depth estimation for visually-guided underwater robots.
\newblock In {\em 2023 IEEE International Conference on Robotics and Automation (ICRA)}, pages 3116--3123. IEEE, 2023.

\bibitem{yuan2022newcrfs}
Weihao Yuan, Xiaodong Gu, Zuozhuo Dai, Siyu Zhu, and Ping Tan.
\newblock Newcrfs: Neural window fully-connected crfs for monocular depth estimation.
\newblock In {\em Proceedings of the IEEE Conference on Computer Vision and Pattern Recognition}, 2022.

\bibitem{yun2024soma}
Seokju Yun, Seunghye Chae, Dongheon Lee, and Youngmin Ro.
\newblock Soma: Singular value decomposed minor components adaptation for domain generalizable representation learning.
\newblock {\em Proceedings of the IEEE/CVF Conference on Computer Vision and Pattern Recognition (CVPR)}, 2025.

\bibitem{zeng2024wordepthvariationallanguageprior}
Ziyao Zeng, Daniel Wang, Fengyu Yang, Hyoungseob Park, Yangchao Wu, Stefano Soatto, Byung-Woo Hong, Dong Lao, and Alex Wong.
\newblock Wordepth: Variational language prior for monocular depth estimation, 2024.

\bibitem{controlnet}
Lvmin Zhang, Anyi Rao, and Maneesh Agrawala.
\newblock Adding conditional control to text-to-image diffusion models.
\newblock In {\em IEEE International Conference on Computer Vision (ICCV)}, 2023.

\bibitem{litemono}
Ning Zhang, Francesco Nex, George Vosselman, and Norman Kerle.
\newblock Lite-mono: A lightweight cnn and transformer architecture for self-supervised monocular depth estimation.
\newblock In {\em Proceedings of the IEEE/CVF Conference on Computer Vision and Pattern Recognition (CVPR)}, pages 18537--18546, June 2023.

\bibitem{adaptive-domain-gen}
Xin Zhang and Ying-Cong Chen.
\newblock Adaptive domain generalization via online disagreement minimization.
\newblock {\em IEEE Transactions on Image Processing}, 32:4247--4258, 2023.

\bibitem{mfuser}
Xin Zhang and Robby~T. Tan.
\newblock Mamba as a bridge: Where vision foundation models meet vision language models for domain-generalized semantic segmentation.
\newblock {\em https://arxiv.org/abs/2504.03193}, 2025.

\bibitem{heap}
Xin Zhang, Jinheng Xie, Yuan Yuan, Michael~Bi Mi, and Robby~T. Tan.
\newblock Heap: Unsupervised object discovery and localization with contrastive grouping.
\newblock {\em arXiv preprint arXiv:2312.17492}, 2024.

\bibitem{zhao2022ITDFA}
Chaoqiang Zhao, Yang Tang, and Qiyu Sun.
\newblock Unsupervised monocular depth estimation in highly complex environments.
\newblock {\em IEEE Transactions on Emerging Topics in Computational Intelligence}, 6(5):1237--1246, 2022.

\bibitem{zhao2022monovit}
Chaoqiang Zhao, Youmin Zhang, Matteo Poggi, Fabio Tosi, Xianda Guo, Zheng Zhu, Guan Huang, Yang Tang, and Stefano Mattoccia.
\newblock Monovit: Self-supervised monocular depth estimation with a vision transformer.
\newblock In {\em 2022 International Conference on 3D Vision (3DV)}, pages 668--678. IEEE, 2022.

\bibitem{VPD}
Wenliang Zhao, Yongming Rao, Zuyan Liu, Benlin Liu, Jie Zhou, and Jiwen Lu.
\newblock Unleashing text-to-image diffusion models for visual perception.
\newblock {\em ICCV}, 2023.

\bibitem{zheng2024physical}
Junhao Zheng, Chenhao Lin, Jiahao Sun, Zhengyu Zhao, Qian Li, and Chao Shen.
\newblock Physical 3d adversarial attacks against monocular depth estimation in autonomous driving.
\newblock In {\em Proceedings of the IEEE/CVF Conference on Computer Vision and Pattern Recognition}, pages 24452--24461, 2024.

\bibitem{23ICRA_Steps}
Yupeng Zheng, Chengliang Zhong, Pengfei Li, Huan-ang Gao, Yuhang Zheng, Bu~Jin, Ling Wang, Hao Zhao, Guyue Zhou, Qichao Zhang, and Dongbin Zhao.
\newblock Steps: Joint self-supervised nighttime image enhancement and depth estimation.
\newblock In {\em 2023 IEEE International Conference on Robotics and Automation (ICRA)}, pages 4916--4923, 2023.

\bibitem{zheng_2020_forkgan}
Ziqiang Zheng, Yang Wu, Xinran Han, and Jianbo Shi.
\newblock Forkgan: Seeing into the rainy night.
\newblock In {\em The IEEE European Conference on Computer Vision (ECCV)}, August 2020.

\bibitem{2017cvpr_egomotion}
Tinghui Zhou, Matthew Brown, Noah Snavely, and David~G. Lowe.
\newblock Unsupervised learning of depth and ego-motion from video.
\newblock In {\em 2017 IEEE Conference on Computer Vision and Pattern Recognition (CVPR)}, pages 6612--6619, 2017.

\bibitem{CycleGAN2017}
Jun-Yan Zhu, Taesung Park, Phillip Isola, and Alexei~A Efros.
\newblock Unpaired image-to-image translation using cycle-consistent adversarial networkss.
\newblock In {\em Computer Vision (ICCV), 2017 IEEE International Conference on}, 2017.

\end{thebibliography}
}

\end{document}